\documentclass[12pt]{article}

\usepackage{spconf}
\usepackage{url,graphicx}
\usepackage[utf8]{inputenc}
\usepackage{babel}
\usepackage{verbatim}
\usepackage{caption}
\usepackage{subcaption}

\usepackage[dvipsnames]{xcolor}
\usepackage{booktabs} % To
\usepackage{multirow}

\newcommand{\mask}{\texttt{MASK}}
\newcommand{\keep}{\texttt{KEEP}}
\newcommand{\rand}{\texttt{RAND}}
\newcommand{\ins}{\texttt{INSERT}}
\newcommand{\drop}{\texttt{DROP}}
\newcommand{\s}{\hspace{0.7mm}}
\title{Warped Language Models for Noise Robust Language Understanding}

\name{Mahdi Namazifar, Gokhan Tur, Dilek Hakkani-T\"ur}

\address{
  Amazon Alexa AI
}

\begin{document} 

\maketitle

\begin{abstract}
 Masked Language Models (MLM) are self-supervised neural networks trained to fill in the blanks in a given sentence with masked tokens. 
 \begin{comment}
by masking (as well as, randomly substituting and leaving intact) some of the tokens of input sentences and minimizing the distance between the modified tokens and their corresponding predicted tokens.
 \end{comment}
  Despite the tremendous success of MLMs for various text based tasks, they are not robust for spoken language understanding, especially for spontaneous conversational speech recognition noise.  In this work we introduce Warped Language Models (WLM) in which input sentences at training time go through the same modifications as in MLM, plus two additional modifications, namely inserting and dropping random tokens. These two modifications extend and contract the sentence in addition to the modifications in MLMs, hence the word ``warped'' in the name. The insertion and drop modification of the input text during training of WLM resemble the types of noise due to Automatic Speech Recognition (ASR) errors, and as a result WLMs are likely to be more robust to ASR noise. Through computational results we show that natural language understanding systems built on top of WLMs perform better compared to those built based on MLMs, especially in the presence of ASR errors. 
\end{abstract}

\section{Introduction}

    Masked Language Models (MLM) and specifically BERT which were introduced in \cite{devlin-etal-2019-bert} are trained by masking some tokens of the input and predicting what the original masked tokens are. 
    The masking of tokens in MLMs is inspired by the Cloze task \cite{doi:10.1177/107769905303000401} in which  participants are asked to guess the missing/masked words in sentences.    Even though  predicting masked tokens, which is referred to as \mask, is the main task in BERT, there are two additional tasks that BERT is trained on. These two tasks are predicting that the token in a given position is kept intact (referred to as \keep), as well as predicting the original token at a specific position in the input sentence that is substituted by another random token (referred to as \rand). Therefore one could think of BERT as a multi-task model.
    A natural consequence of thinking of BERT as a multi-task model is the idea of adding even more tasks to the list of tasks that BERT is trained to solve in order to achieve a richer language model with a more diverse capabilities, or even making the process of training these language models more efficient. In this work we introduce two additional tasks to the set of tasks that BERT is trained on, namely we
    \begin{itemize}
        \item insert a random token in a random position in the input sentence and train the model to predict a special token for those positions, which we refer to as \ins.
        
        \item delete the token at a random position in the input sentence and train the model to predict what the deleted token is, which we refer to as \drop.
    \end{itemize}

\begin{figure}[h]
    \centering
    \fbox{\includegraphics[width=0.48\textwidth]{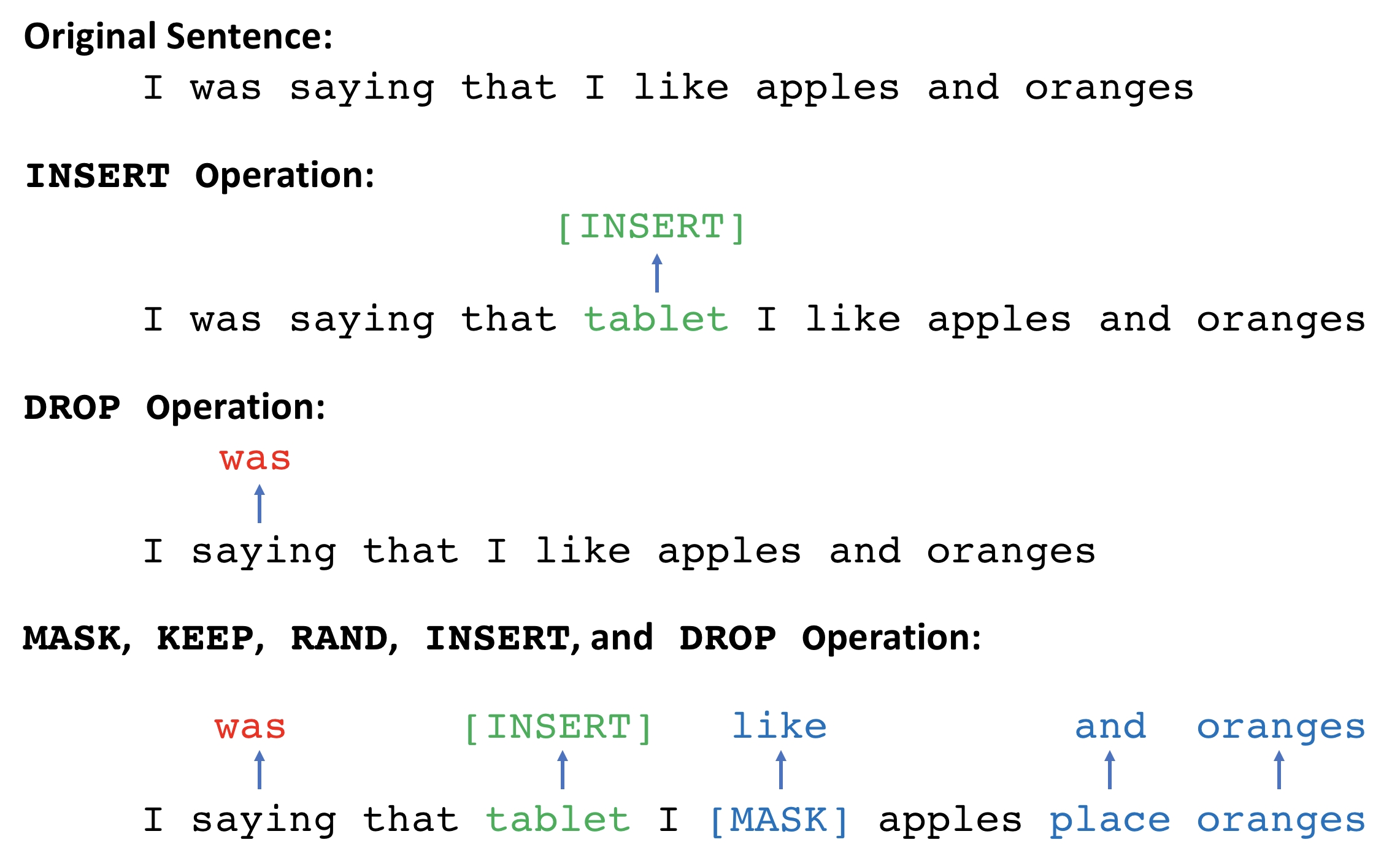}}
    \caption{\footnotesize Example of warping in WLMs. The inserted tokens get the special insert token as their label (shown in green). The tokens following a deleted token get the deleted token as their label (shown in red). The \mask, \keep, and \rand \s\s operations from original BERT are shown in blue.}
    \label{fig:warp}
\end{figure}

    We refer to these new language models as Warped Language Models (WLM) since in the training process the input text is first ``warped'' (masked, extended, contracted, etc.) through the operations mentioned above, namely some tokens are masked, some tokens are substituted by other random tokens, some tokens are deleted, and some random tokens are inserted in random positions.
    
    \begin{comment}
    %%% NOTE!!!
    \textcolor{red}{[For the extended version of the paper we need to have a broader introduction into this approach of introducing noise, instead of jumping right into SLU]}
    \end{comment}
    The noise that is introduced by these warping operations, among other things, forces the language model to be more robust to input noise. 
    As BERT and WLM are based on transformers \cite{NIPS2017_7181} which themselves heavily rely on self-attention, they pay attention to the entire input even in the presence of noise and errors.
    Such input noise is specifically inherent in Spoken Language Understanding (SLU) tasks that as input use the output of an Automatic Speech Recognition (ASR) system. ASR noise is quite similar to the noise that is introduced by the warping operations \rand, \ins, and \drop \s in nature. More specifically, the ASR output might have some words of the spoken sentences substituted by other words, some words that are inserted into the spoken sentence, or some words of the spoken sentence that are deleted.  In this work we focus on the impact of the additional two tasks that we introduce (\ins and \drop) on SLU systems built on top of WLMs. 
    
    In our experiments through evaluation of Spoken Language Understanding (SLU) systems trained on the Airline Travel Information System (ATIS) data set \cite{hemphill-etal-1990-atis, dahl-etal-1994-expanding, 5700816} we show that SLU systems built on top of WLMs are more robust to ASR noise compared to those built on top of MLMs, and specifically BERT.

\section{Related Work}
Following the success of BERT, GPT-2 \cite{radford2019language}, and ELMo \cite{Peters:2018}, that used masking or next token prediction as training tasks, a number of other works have looked into different tasks as basis for training language models. For instance instead of masking the inputs ELECTRA \cite{clark2020electra} substitutes some of the tokens with tokens sampled from a generator network and a discriminator detects what token was substituted. In BART \cite{lewis-etal-2020-bart} the authors train a denoising auto-encoder where the noising function corrupts the input through token masking, token deletion, sentence span masking, sentence permutation and document rotation. Token deletion in our WLM is different from token deletion in BART since in BART the model decides which positions (i.e., word boundaries) are missing inputs whereas in WLM the model predicts the missing token for the token that comes after it. In another recent work \cite{xu2020mcbert} the task based on which the language model is trained is a multi-choice Cloze task with ``reject'' as one of the choices. In this work a separate network provides training input and choices. 

Regarding the applications of WLMs that are of interest in this work, namely SLU from ASR output, several recent works have looked into creating models that are robust to ASR noise. In \cite{wang-etal-2020-data} the authors use an ASR simulator to inject noise in the otherwise error-free training data for SLU tasks. In a different yet related approach in \cite{9054689} a ``confusion'' loss is added to the language model's training loss to force it to produce similar embeddings for acoustically confusable words. In \cite{9053213} the authors propose a joint model to both correct ASR output and language understanding using the ASR n-best hypotheses. There are also several works that use transformer \cite{NIPS2017_7181} based language models to improve ASR output quality. \cite{9053051} trains a transformer based encoder-decoder architecture to translate ASR output into semantically and grammatically correct text. In \cite{Irie2019LanguageMW} the authors use transformer-based language models for lattice rescoring \cite{Sundermeyer2014LatticeDA}.

\section{Warped Language Models}
As we described in the Introduction section, we define WLMs to train on two additional task on top of the \mask, \keep, and \rand\s tasks that are used in BERT. These two additional tasks are \ins\s and \drop. In the next two subsections we describe the details of these two tasks and Figure~\ref{fig:warp} depicts examples of each task.
\subsection{\ins}
In this task random tokens are inserted at random positions in the input text. For these positions the language model is trained to predict a special token that is only used for indicating an inserted token. In other words during training the inserted tokens get the \ins \s special token as label. The embedding for this special token is randomly initialized at the beginning of the training process, and since it does not appear in the training data, there is no back-propagation into its embeddings and as a result, this embedding remains intact throughout the training process, while all other token embeddings continue to change by the back-propagation process.

\subsection{\drop}
In the \drop \s task, during the training of the WLM, some tokens are randomly deleted from the input sentences. For each deleted token the model is forced to predict the deleted token for its subsequent token. In other words, during training of the WLM, the token following a deleted token gets the deleted token as its label.
% NOTE 4 Mahdi: If you have implementations of variations of the drop task, I think they would be good to include, as I am quite sure reviewers would ask why not word boundaries?

\subsection{Masking vs Warping}
The operations that are applied on the input task during training of MLMs, namely \mask, \keep, and \rand \s do not change the length or the structure of the input sentences. These operations could simply be implemented using element-wise operations of an operating vector (a vector that indicates where to \mask, \keep, or \rand). One could think of this operating vector applied element-wise on the input tokens vector as performing ``masking'' on the input, as it is a layer that sits on top of the input vector. On the other hand, the two additional operations that we introduce in WLMs, namely \ins\s and \drop \s change the length and structure of the input sentence. These two operations cannot be implemented by simply applying an operating vector to the vector of the input tokens, as is the case for the operations in MLMs. 

The two new operations \ins\s and \drop \s are fundamentally different from the original operations of MLMs (\mask, \keep, \rand). The operation \ins \s forces the model to return a fixed token every time it determines that a token was inserted into the input sentence, given the context. The operation \drop \s is the more bizarre of the two since it anchors the sentence at the positions followed by a deleted token to the previous position by labeling the token after a deleted token with the deleted token. In all of the other operations in WLMs, the label at a position is directly indicating what the token in that position should be (in the case of \ins \s the token should be none), but in the \drop \s operation the label is indicating what the previous token should be. Intuitively this operation might seem destructive to the training process, but as we show in the experimental results, the combination of the two new operations not only does not impair the convergence of the language model, it in fact improves the performance of the language model on downstream SLU tasks involving ASR noise. 

Finally it should also be noted that WLMs are generalization of MLMs where the \ins \s and \drop \s are introduced in addition to \mask, \keep, and \rand operations at training time. Note that the difference between WLMs and MLMs are only in how they are trained and beyond that, the architecture, input and output formats, etc. are identical.

\subsection{Illegal Operations}
Another way that the two operations that we introduce in this paper (\ins, \drop) are different from MLM's original operations (\mask, \keep, \rand) is that the these new operations could create warping of sentences that should not be permitted. The first case that creates an illegal warping is where a \mask, \keep, or \rand \s operation appears immediately after a \drop. 
% NOTE 4 Mahdi: Shall we include a figure here for easy explanation?
This is an illegal case because if it happens the token after the deleted token would get two different labels, one of which would be the deleted token for the \drop \s operation and the other one would be the label associated with the \mask, \keep, or \rand \s operation. The second case that creates an illegal warping is when more than one consecutive \drop \s operations happen. In this case there would be no way to indicate the first deleted token as a label. In the implementation of WLMs we make sure that these two illegal cases do not occur during the training process.

\section{Experiments}
We implemented WLM by generalizing the BERT implementation within the XLM framework \cite{lample2019cross} to cover \ins \s and \drop\s operations. We then train a BERT model as well as a WLM model for English using the entire Wikipedia data from scratch, to use as a baseline. For these models we used 512 for the embedding dimensions and a 12 layer transformer encoder module with 16 attention heads. In this setup the model has about 56 million parameters. We trained these models (WLM and BERT) using 8 GPUs on an Amazon AWS p3.16xlarge instance for 100 epochs. The training process takes over 4 days. In the original BERT paper the authors propose to first randomly pick 15\% of tokens, and then among those tokens \mask\s 80\%, \keep\s 10\%, and \rand \s 10\% of them. We follow this exact split for training BERT. However for training WLM, we first randomly pick 15\% of tokens (similar to BERT), but then among those tokens we \mask\s 60\%, \keep\s 10\%, \rand\s 10\%, \ins\s 10\%, and \drop \s 10\%. 

Table \ref{table:1} shows perplexity and accuracy of WLM and BERT after 100 epochs of training measured on the same validation and test sets. 
% NOTE 4 Mahdi: Are the accuracies comparable? These are tackling two different tasks, it would be good to note that these are not exactly comparable
% NOTE 4 Mahdi: Another related question is: Can we create a test set that is comparable? i.e., what if we use BERT test set for WLM or some randomly distorted test set for both and list accuracy for each sub-task or their confusion matrix? i.e., insertion, deletion, etc. and how BERT would treat these as other types of errors.
From these numbers we can see that the final BERT model (after 10 epochs of training) has a slightly better perplexity and accuracy on the test sets.
Note the the architecture and the number of parameters of both models are identical.
\begin{comment}
\begin{figure*}%
    \centering
    \subfloat[]{{\includegraphics[width=7cm]{perplexity.png} }}%
    \qquad
    \subfloat[]{{\includegraphics[width=7cm]{accuracy.png} }}%
    \caption{\footnotesize Left: Validation perplexity of WLM and BERT throughout training. Note that the Y axis is in log scale. Right: Validation accuracy of WLM and BERT throughout training. Both WLM and BERT models have around 56 million parameters.}%
    \label{fig:1}%
\end{figure*}

\end{comment}

\begin{table}[h!]
\centering
\begin{tabular}{||c | c | c|| c | c ||} 
 \hline
  &  \multicolumn{2}{c||}{Validation} & \multicolumn{2}{|c||}{Test}\\
  [0.5ex] 
  \hline
  & Perp. &  Acc. & Perp. & Acc.\\ [0.5ex] 
 \hline\hline

 WLM (56M) & 7.68 & 59.61 & 7.80 & 59.01\\ [0.5ex]
 BERT (56M) & 7.34 & 60.22 & 7.43 & 59.72 \\ [0.5ex]
 \hline
\end{tabular}
\caption{\footnotesize Perplexity and Accuracy of  WLM and BERT after 100 epochs of training. Both models have approximately 56 million parameters. Accuracy numbers are computed identically for WLM and BERT}
\label{table:1}
\end{table}

\subsection{WLM for SLU with ASR Noise}

Since WLM encounters \ins \s and \drop \s operations during its training we expect that it would be more robust to these types of noise in the input text. These types of noise are frequently seen in ASR outputs. In this section through our experiments we show that SLU systems built for ASR outputs that use WLM perform better than the same models that use BERT instead of WLM.

\subsubsection{Data: ATIS with ASR Noise}
The ATIS dataset \cite{hemphill-etal-1990-atis, dahl-etal-1994-expanding, 5700816} is a widely used  benchmark for natural language understanding and provides audio recordings as well as manual transcriptions of utterances inquiring a flight information system. The ATIS dataset also provides manual labels for the 81 slots and 18 intents that appear in the data. In this paper, we use the ASR output for this dataset that was released by Huang and Chen~\cite{9054689}. We aligned these ASR outputs with the manual transcripts using SCTK, the NIST scoring toolkit~\footnote{https://github.com/usnistgov/SCTK}. We obtained the slot tags for the ASR output by transferring the labels from manual transcriptions to aligned words in the ASR output and updating the labels to preserve the IOB-style annotation \cite{ramshaw-marcus-1995-text}.

The original ATIS dataset release does not contain a validation set, hence, we used the training, validation, and test set split from \cite{zhang2019joint}, which includes 4478, 500, and 893 utterances in the training, validation and test subsets, respectively. Table~\ref{table:wer} shows the word error rate (WER) as well as insertion (INS), deletion (DEL), and substitution (SUBS) error rates in the training, validation, and test subsets.

\begin{table}[h!]
\centering
\begin{tabular}{||l | c || c| c | c ||} 
 \hline
  Subset & WER& INS& DEL & SUBS \\
  \hline
Train+Val & 18.6\% & 3.3\% & 2.4\% & 12.9\% \\
Test & 15.9\% & 2.9\% & 1.5\% & 11.5\% \\
 \hline
\end{tabular}
\caption{\footnotesize Word error rate (WER), insertion (INS), deletion (DEL), and substitution (SUBS) error rates in the training, validation, and test subsets of the ATIS ASR outptus released by~\cite{9054689}.}
\label{table:wer}
\end{table}

%\textcolor{red}{Dilek: add ASR WER and insert/del/subs rates.}

\begingroup
\setlength{\tabcolsep}{4mm} % Default value: 6pt
\renewcommand{\arraystretch}{1.5} % Default value: 1

\begin{table*}[ht!]
\footnotesize
\centering
\begin{tabular}{|c||c|c|c||c|c|c||c|c|c||}
\hline
%\vspace{-10}

\multirow{2}{0mm}{} & \multicolumn{3}{c||}{Train on Transcribed} & \multicolumn{3}{c||}{Train on Transcribed } & \multicolumn{3}{c||}{Train on ASR} \\\\[-5ex]
& \multicolumn{3}{c||}{Test on Transcribed} & \multicolumn{3}{c||}{Test on ASR} & \multicolumn{3}{c||}{Test on ASR} \\
\hline
  & intent &	slot & joint & intent &	slot & joint & intent &	slot & joint \\[0.5ex]
\hline

 WLM (56M) & 97.91 & 93.65 & 84.23 & 93.93 & \bf{84.14} & \bf{59.59} & 95.51 & \bf{88.72} & \bf{70.10} \\[0.5ex]
%\midrule
 BERT (56M) & 97.77 &	93.41 &	83.18 & 93.69 &	83.08 &	58.25 & 95.28 &	87.39 & 	67.89 \\[0.5ex]
 \hline
\end{tabular}
\caption{\footnotesize Performance of the SLU model built on top of WLM and BERT, where both WLM and BERT models have around 56 million parameters. Values for ``intent'' are accuracy of intent detection, values for ``slot'' are F1 measure of slot detection, and, finally, values for ``joint'' are joint accuracy which is the percentage of utterances which have all slots and intents correctly annotated. Bold values represent statistically significant difference at p-value of 0.05.}
\label{table:2}
\end{table*}
\endgroup
\vspace{-11mm}
\subsubsection{SLU Model}
\begin{figure}[h]
    \centering
    \includegraphics[width=0.38\textwidth]{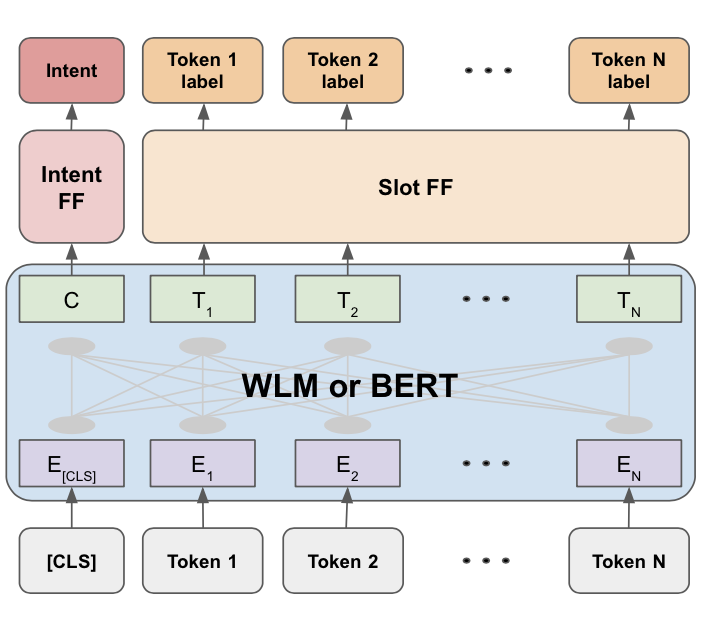}
    \caption{\footnotesize SLU model following the architecture proposed in \cite{DBLP:journals/corr/abs-1902-10909}. ``FF'' in the boxes above the BERT box stands for feed forward network.}
    \label{fig:nlu}
\end{figure}

We build our SLU models with WLM and BERT following the general architecture proposed in  \cite{DBLP:journals/corr/abs-1902-10909} to jointly predict intents and slots on top of WLM and BERT as shown in Figure \ref{fig:nlu}. More specifically we add an intent prediction head and slot prediction heads to the output of WLM and BERT. We use the ATIS training set to train this model by back-propagating the loss through the entire model (intent classification head, slot tagging heads, and the language model).
% NOTE 4 Mahdi: Can we include a figure for model architecture as in the Alibaba paper?

\subsubsection{Evaluation}
For reporting performance, we use three measures: intent accuracy (the percentage of  utterances with correct intents), slot F1 score (CoNLL Style, \cite{tjong-kim-sang-de-meulder-2003-introduction}), and sentence level joint accuracy (the percentage of utterances which have all the slots and intents accurately annotated).

\subsubsection{Experimentation Results}
In our experiments, as a result of our aligning transcribed and ASR data, for each of the training and test sets we have the intents and slots labels for both transcribed and ASR data. Therefore we use the following three settings for our experiment:

\begin{itemize}
    \item Train on transcribed data and test on transcribed data (Transcribed--Transcribed)
    \item Train on transcribed data and test on ASR data (Transcribed--ASR)
    \item Train on ASR data and test on ASR data (ASR--ASR)
\end{itemize}

For each of these settings we run the training process 5 times and the numbers that we report are averaged over these 5 runs. Table \ref{table:2} summarizes the results of these experiments. Numbers in bold show the values that are higher in a statistically significant manner at p-value of 0.05.
From the table we could see that the difference between intent accuracy for WLM and BERT in all of the three settings is negligible. This could be due to the fact that ASR noise does not interfere with intent detection significantly since, intuitively, even in presence of some noise, detecting the intent of an utterance is still possible. Also the difference between WLM and BERT for slot detection in the case where both training and testing is done on transcribed data is notably negligible. The differences between slot detection F1 scores and joint accuracy measures become more prominent when we consider the cases where both training and testing or only testing is done on the ASR data. In both these cases it is clear from the table that the models perform significantly worse overall due to the presence of ASR noise in either the test set only or in both training and test sets. This drop in performance is more pronounced for the case where training is done on transcribed data and testing is done on ASR data, which could be explained by distribution mismatch between training and test sets. In this case, we see that WLM improves the performance on slot detection and joint accuracy significantly. This improvement is also clear in the case where both training and testing is done on the ASR data, where we see that the joint accuracy of the BERT based model is 67.89, whereas for the WLM based model it is 70.10.

 For the case where both training and testing is done on the transcribed data the slot F1 measure (93.65) is lower than the numbers reported for BERT-base in  \cite{DBLP:journals/corr/abs-1902-10909} (95.59). But note that BERT-base model has around 110 million parameters, but we trained our BERT and WLM with 56 million parameters due to the long time it takes to train these models (over 4 days on an 8 GPU machine).

    \begin{comment}
    \begin{itemize}
    %%% NOTE!!!
    \item \textcolor{red}{Implementation details}
    \item \textcolor{red}{Complexities of INSERT and DROP (illegal changes, ...)}
    \item Our notation i.e. 61111 and 81100
    \end{itemize}
    \end{comment}

\begin{comment}
%%% NOTE!!!
\textcolor{red}{
\subsection{WLM vs MLM}
\begin{itemize}
    \item Training process:
    \begin{itemize}
        \item Validation perplexity and accuracy during training
        \item Validation accuracy with noisy input
    \end{itemize}
\end{itemize}
}

\end{comment}

\begin{comment}
%%% NOTE!!!
\textcolor{red}{
\subsection{WLM vs MLM on GLUE}
\begin{itemize}
    \item Experimental results
\end{itemize}
}
\end{comment}

\section{Conclusions}
In this work we introduce WLMs, that are generalization of MLMs, through introduction of two additional training tasks, namely \ins \s and \drop \s where the model is trained to predict where a random token is inserted into or deleted from the input sequence during training. Due to the new noise that WLMs see at training time, they are more robust to noisy input. We demonstrate this by showing that SLU models built on WLMs perform better than the same models built on MLMs when ASR noise is present.  

There is however much more to be studied regarding WLMs to better understand their strengths and weaknesses. Moreover, there are other behaviors of WLMs that are not understood to us. For instance the fact that in the \drop\s operation the label of tokens come from the previous position does not throw off the training process needs to be studied in more depth. Moreover, the fact that there is no distinguishing factor for the model between the label of a token following a \drop ed token and the label of a \rand \s token, and yet somewhat surprisingly the training process still converges, needs to be studied more. 
\begin{comment}

On the topic of different types of modifications of input sentences at training time a research question here is what are other tasks that could be added to the list of the tasks that WLMs try to solve and of what of different types could they be.
%%% NOTE!!!
\begin{itemize}

    \item \textcolor{red}{Why does DROP doen't completely throw off the learning process}
    \item \textcolor{red}{Why does WLMs start converging much faster than MLMs? How could this be exploited?}
    \item \textcolor{red}{Other losses?}

    \item SLU implications
\end{itemize}
\end{comment}

\bibliographystyle{IEEEbib}
%\bibliography{main}

\end{document}